\documentclass[preprint,12pt]{elsarticle}

\usepackage[utf8]{inputenc}
\usepackage{amsmath,amssymb,amsfonts}
\usepackage{graphicx}
\usepackage{booktabs}
\usepackage{hyperref}
\usepackage{xcolor}
\usepackage{algorithm}
\usepackage{algorithmic}
\usepackage{multirow}
\usepackage{caption}
\usepackage{subcaption}

\graphicspath{{figures/}}

\makeatletter
\def\ps@pprintTitle{%
  \let\@oddhead\@empty
  \let\@evenhead\@empty
  \def\@oddfoot{\reset@font\hfil\thepage\hfil}
  \let\@evenfoot\@oddfoot
}
\makeatother

\begin{document}

\begin{frontmatter}

\title{Default Machine Learning Hyperparameters Do Not Provide Informative Initialization for Bayesian Optimization}

\author[comillas]{Nicol\'as Villagr\'an Prieto}
\author[comillas,iit]{Eduardo C. Garrido-Merch\'an\corref{cor1}}
\cortext[cor1]{Corresponding author}
\ead{ecgarrido@comillas.edu}

\affiliation[comillas]{organization={Department of Telematics and Computing, Universidad Pontificia Comillas},
            city={Madrid},
            country={Spain}}
\affiliation[iit]{organization={Institute for Research in Technology (IIT), Universidad Pontificia Comillas},
            city={Madrid},
            country={Spain}}

\begin{abstract}
Bayesian Optimization (BO) has become a standard tool for hyperparameter tuning in machine learning due to its sample efficiency when exploring expensive black-box functions. While most BO pipelines begin with uniform random initialization, default hyperparameter values shipped with popular ML libraries such as scikit-learn encode implicit expert knowledge and could, in principle, serve as informative starting points that accelerate convergence. This hypothesis, despite its intuitive appeal, has remained largely unexamined. In this work we formalize the idea by initializing BO with points drawn from truncated Gaussian distributions centered at library defaults and compare the resulting optimization trajectories against a standard uniform-random baseline. We conduct an extensive empirical evaluation spanning three BO back-ends (BoTorch, Optuna, Scikit-Optimize), three model families (Random Forests, Support Vector Machines, Multilayer Perceptrons), and five benchmark datasets covering both classification and regression tasks. Performance is assessed through convergence speed and final predictive quality, and statistical significance is determined via one-sided binomial tests. Across all experimental conditions, default-informed initialization yields no statistically significant advantage over purely random sampling, with $p$-values ranging from 0.141 to 0.908. A complementary sensitivity analysis on the prior variance confirms that, while tighter concentration around the defaults improves early evaluations, this transient benefit vanishes as the optimization progresses, leaving final performance unchanged. Our results provide no evidence that default hyperparameters encode useful directional information for optimization. We therefore recommend that practitioners treat hyperparameter tuning as an integral part of model development and favor principled, data-driven search strategies over heuristic reliance on library defaults.
\end{abstract}

\begin{keyword}
Bayesian Optimization \sep Hyperparameter Tuning \sep Initialization Strategies \sep Default Hyperparameters \sep Gaussian Processes \sep Machine Learning
\end{keyword}

\end{frontmatter}

\section{Introduction}
\label{sec:intro}

Bayesian optimization is the state-of-the-art framework for optimizing expensive-to-evaluate black-box functions whose analytical form is unavailable and whose gradient information cannot be obtained~\cite{garnett2023bayesian,frazier2018tutorial}. The canonical setting builds a probabilistic surrogate---typically a Gaussian Process (GP)---over the objective landscape and sequentially selects evaluation points by maximizing an acquisition function that balances exploration of uncertain regions with exploitation of promising ones~\cite{rasmussen2006gaussian}. Because each evaluation may involve training a full machine-learning pipeline, the sample efficiency of BO has made it the method of choice for hyperparameter tuning~\cite{snoek2012practical,turner2021bayesian}.

A critical yet under-examined component of every BO pipeline is its initialization strategy. The prevailing practice draws a small set of initial points uniformly at random from the search domain, implicitly assuming that no prior knowledge about the location of high-performing configurations is available. From a Bayesian standpoint, however, this amounts to placing a non-informative prior on the objective function, discarding any domain expertise that might accelerate the search. In practice, widely used ML libraries such as \texttt{scikit-learn}~\cite{pedregosa2011scikit}, \texttt{XGBoost}, and \texttt{LightGBM} ship with carefully chosen default hyperparameter values. These defaults have been selected by library developers through extensive empirical testing to perform broadly well across a wide variety of datasets and tasks. One might therefore conjecture that they encapsulate meaningful prior information about where good hyperparameter configurations tend to lie, and that incorporating this information into the BO initialization could yield faster convergence and better final solutions.

This conjecture is not merely academic. If default hyperparameters do encode informative structure about the loss landscape, centering the initial sampling distribution at the default values would represent a zero-cost enhancement to any existing BO pipeline, requiring no additional data, no meta-learning infrastructure, and no modification of the surrogate or acquisition function. Conversely, if defaults turn out to be uninformative for optimization, the finding would carry important practical implications: it would warn practitioners against the widespread but untested assumption that ``good enough out of the box'' translates into ``useful starting point for search,'' and it would reinforce the need for principled, uncertainty-aware exploration from the outset.

\paragraph{Terminology} Throughout this paper we use the term ``prior'' in the broad Bayesian sense of prior belief about where high-performing configurations lie, as encoded through the initialization strategy. This should not be confused with the GP prior (mean function, kernel hyperparameters), which we do not modify. Specifically, we study whether centering the \emph{initialization distribution} of BO around library defaults---rather than using uniform random sampling---translates into measurable optimization gains.

In this paper we subject this hypothesis to rigorous empirical scrutiny. We design a controlled experimental framework in which default-informed initialization---realized through truncated Gaussian sampling centered at the library default values---is compared head-to-head against standard uniform random initialization across a combinatorial grid of optimization libraries, machine-learning models, and benchmark datasets. The evaluation protocol measures both convergence speed and final predictive quality, and statistical significance is assessed via one-sided binomial tests that directly quantify the probability of default-based strategies outperforming the random baseline beyond chance. We further complement the main analysis with a sensitivity study that varies the concentration of the prior around the defaults, thereby probing whether the degree of informativeness of the initialization modulates the outcome.

Our results are unequivocal. Across all experimental conditions---three BO back-ends (BoTorch, Optuna, Scikit-Optimize), three model families (Random Forests, Support Vector Machines, Multilayer Perceptrons), five datasets, and multiple prior-variance settings---default-informed initialization provides no statistically significant advantage over random sampling. The binomial $p$-values range from 0.141 to 0.908, far above any conventional significance threshold. The sensitivity analysis further reveals that while tighter concentration around defaults temporarily inflates early-iteration performance (because defaults are engineered to be safe single-shot configurations), this transient head start is rapidly erased as BO explores the space, and the final solution quality remains indistinguishable across all prior widths. We find no evidence that default hyperparameters are informative for optimization: they are a shortcut for deployment, not a substitute for principled search.

The remainder of this paper is organized as follows. Section~\ref{sec:background} introduces the mathematical foundations of Gaussian Processes and Bayesian optimization. Section~\ref{sec:related} reviews the state of the art, with emphasis on prior modeling and initialization strategies. Section~\ref{sec:method} presents the research hypotheses and our methodology for testing them. Section~\ref{sec:experiments} details the experimental setup. Section~\ref{sec:results} presents and discusses the results, including the sensitivity analysis. Finally, Section~\ref{sec:conclusion} draws conclusions and outlines directions for future work.

\section{Background}
\label{sec:background}

This section introduces the mathematical foundations underlying Bayesian optimization, covering Gaussian Process regression, the construction of acquisition functions, and the complete BO loop. These concepts are essential for understanding how initialization---the central object of study in this work---enters and shapes the optimization process.

A Gaussian Process (GP) is a collection of random variables, any finite subset of which follows a joint Gaussian distribution~\cite{rasmussen2006gaussian}. A GP is fully specified by a mean function $m(x) = \mathbb{E}[f(x)]$ and a covariance (kernel) function $k(x, x') = \mathbb{E}[(f(x) - m(x))(f(x') - m(x'))]$, so that $f(x) \sim \mathcal{GP}(m(x), k(x, x'))$. For a finite set of inputs $X = \{x_1, \ldots, x_n\}$, the corresponding function values $\mathbf{f} = [f(x_1), \ldots, f(x_n)]^\top$ are jointly normally distributed as $\mathbf{f} \sim \mathcal{N}(\mathbf{m}, K)$, where $\mathbf{m} = [m(x_1), \ldots, m(x_n)]^\top$ and $K_{ij} = k(x_i, x_j)$.

The kernel function controls the inductive bias of the GP by encoding assumptions about the smoothness and correlation structure of the objective function. A widely used choice is the squared exponential (SE) kernel, $k(x, x') = \sigma_f^2 \exp\!\big(-\|x - x'\|^2 / (2\ell^2)\big)$, where $\sigma_f^2$ is the signal variance and $\ell$ is the length-scale. This kernel assigns high covariance to nearby points and low covariance to distant ones. Crucially for the present study, the kernel governs how information propagates from observed evaluations to unexplored regions: if initial observations are clustered around a single point (e.g., the library default), the GP posterior will have low predictive variance near that cluster but remain highly uncertain elsewhere, potentially suppressing exploration of distant regions where the true optimum may lie.

Given a dataset of $t$ observations $\mathcal{D}_t = \{(x_i, y_i)\}_{i=1}^{t}$ with $y_i = f(x_i) + \varepsilon$, $\varepsilon \sim \mathcal{N}(0, \sigma_n^2)$, the GP posterior at an arbitrary test point $x$ is Gaussian with mean and variance
\begin{equation}
\mu_t(x) = k(x, X)^\top [K + \sigma_n^2 I]^{-1} \mathbf{y}, \qquad
\sigma_t^2(x) = k(x, x) - k(x, X)^\top [K + \sigma_n^2 I]^{-1} k(x, X),
\label{eq:gp_posterior}
\end{equation}
where $k(x, X)$ is the vector of covariances between $x$ and the training inputs. The posterior mean $\mu_t(x)$ serves as the surrogate model's prediction, and the posterior variance $\sigma_t^2(x)$ quantifies predictive uncertainty. Together, they provide the ingredients for principled sequential decision-making.

Bayesian optimization exploits the GP surrogate to select the next evaluation point via an acquisition function $\alpha(x)$ that balances exploration (querying regions of high uncertainty) with exploitation (querying regions of high predicted performance). A widely used acquisition function, and the one adopted in this study, is Expected Improvement (EI)~\cite{snoek2012practical}:
\begin{equation}
\alpha_{\mathrm{EI}}(x) = \mathbb{E}\!\Big[\max\!\big(f(x) - f(x^{+}),\; 0\big)\Big] = \big(\mu_t(x) - f(x^{+})\big)\,\Phi(z) + \sigma_t(x)\,\phi(z),
\label{eq:ei}
\end{equation}
where $f(x^{+}) = \max_{i \leq t} y_i$ is the best value observed so far, $z = (\mu_t(x) - f(x^{+})) / \sigma_t(x)$, and $\Phi(\cdot)$, $\phi(\cdot)$ are the standard normal CDF and PDF respectively. EI naturally balances the two terms: the first rewards points with high predicted mean (exploitation), while the second rewards points with high uncertainty (exploration). The next evaluation is placed at $x_{t+1} = \arg\max_x \alpha_{\mathrm{EI}}(x)$.

\begin{figure}[t]
\centering
\includegraphics[width=0.85\textwidth]{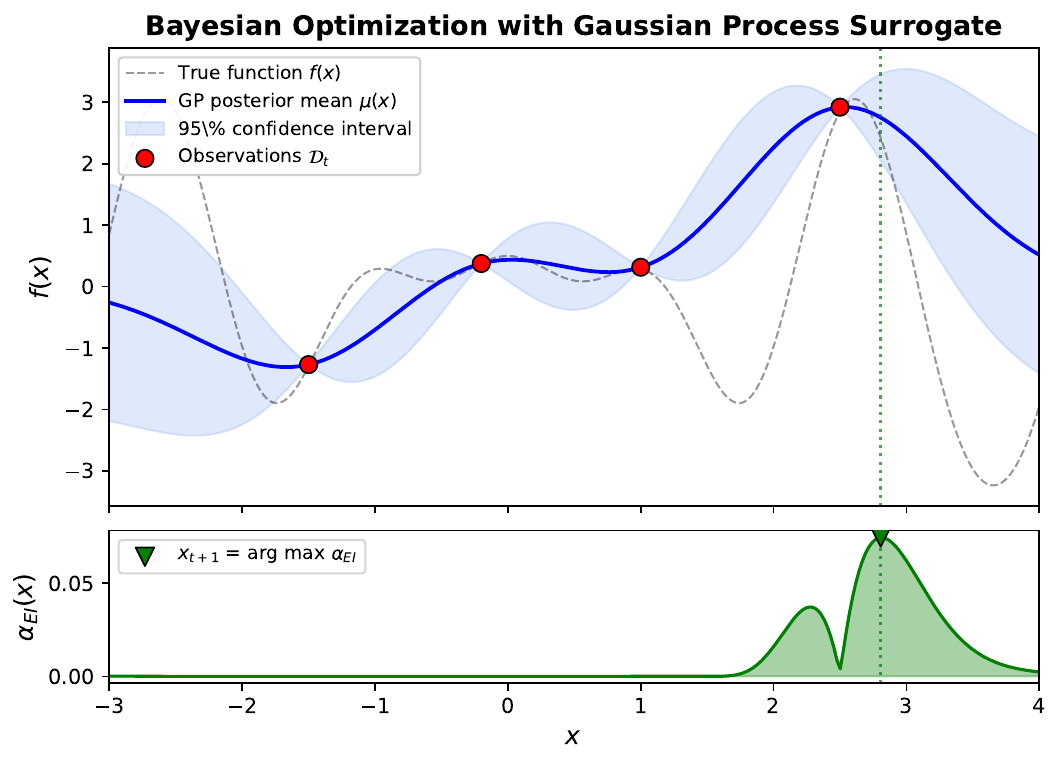}
\caption{Illustration of one iteration of Bayesian optimization. \textbf{Top:} The GP posterior (blue curve and shaded 95\% confidence interval) is fitted to four observations (red dots). The dashed line shows the true (hidden) objective. \textbf{Bottom:} The Expected Improvement acquisition function identifies the next evaluation point $x_{t+1}$ (green marker) by balancing exploration and exploitation.}
\label{fig:bo_illustration}
\end{figure}

The complete BO procedure is summarized in Algorithm~\ref{alg:bo}. The algorithm starts from an initial dataset $\mathcal{D}_0$---whose construction is precisely the focus of this paper---and iteratively refines its surrogate model, selecting new query points until the evaluation budget $T$ is exhausted. Figure~\ref{fig:bo_illustration} illustrates a single iteration of this loop, showing the GP posterior, its uncertainty band, and the acquisition function that selects the next evaluation point.

\begin{algorithm}[t]
\caption{Bayesian Optimization}
\label{alg:bo}
\begin{algorithmic}[1]
\REQUIRE Search domain $X \subset \mathbb{R}^d$, evaluation budget $T$, acquisition function $\alpha$
\REQUIRE Initial dataset $\mathcal{D}_0 = \{(x_i, y_i)\}_{i=1}^{n_0}$ \COMMENT{Key design choice studied in this work}
\FOR{$t = n_0, n_0 + 1, \ldots, T$}
    \STATE Fit GP surrogate to $\mathcal{D}_t$: compute $\mu_t(x)$ and $\sigma_t^2(x)$ via Eq.~(\ref{eq:gp_posterior})
    \STATE Select next query point: $x_{t+1} = \arg\max_{x \in X} \alpha(x \mid \mu_t, \sigma_t)$
    \STATE Evaluate the objective: $y_{t+1} = f(x_{t+1}) + \varepsilon$
    \STATE Augment the dataset: $\mathcal{D}_{t+1} = \mathcal{D}_t \cup \{(x_{t+1}, y_{t+1})\}$
\ENDFOR
\RETURN $x^* = \arg\max_{(x,y) \in \mathcal{D}_T} y$
\end{algorithmic}
\end{algorithm}

\begin{figure}[t]
\centering
\includegraphics[width=\textwidth]{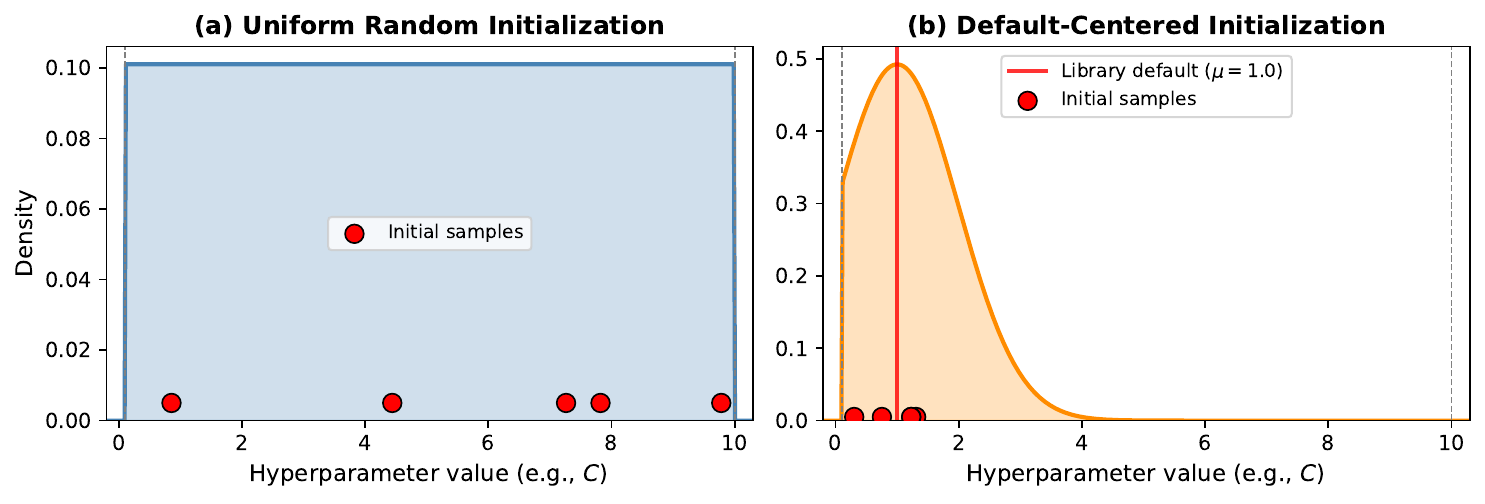}
\caption{Comparison of initialization strategies for Bayesian optimization. \textbf{(a)}~Uniform random initialization draws points (red dots) from a uniform distribution over the entire hyperparameter domain $[a, b]$. \textbf{(b)}~Default-centered initialization draws points from a truncated Gaussian centered at the library default value $\mu$, concentrating initial evaluations in the neighborhood of the default configuration.}
\label{fig:prior_comparison}
\end{figure}

The choice of the initial dataset $\mathcal{D}_0$ in Line~1 of Algorithm~\ref{alg:bo} is the critical design decision investigated in this paper. The standard approach samples $\mathcal{D}_0$ uniformly at random from $X$, which provides broad coverage but ignores any available domain knowledge. Our proposal is to instead draw $\mathcal{D}_0$ from a truncated Gaussian distribution centered at the default hyperparameter values provided by ML libraries, encoding the belief that the optimum lies near the defaults. Figure~\ref{fig:prior_comparison} visually contrasts these two initialization strategies: the uniform approach spreads initial evaluations across the entire domain, while the default-centered approach concentrates them near the library default. The question we address is whether the latter strategy provides any measurable advantage.

\begin{figure}[t]
\centering
\includegraphics[width=\textwidth]{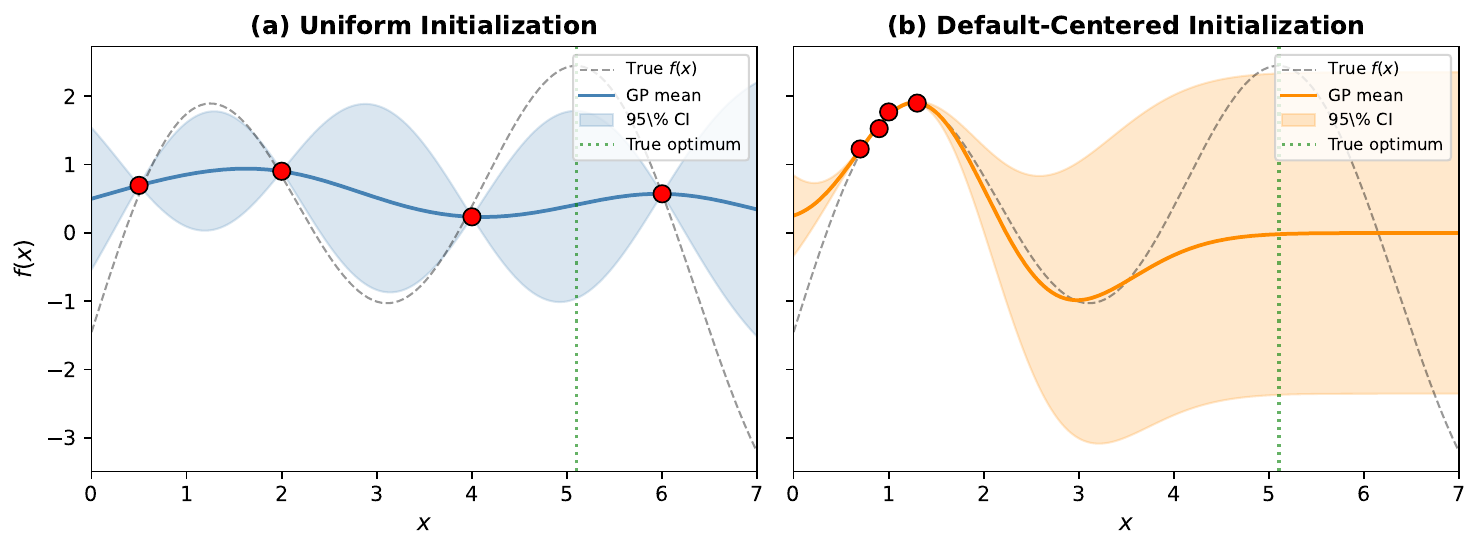}
\caption{Effect of initialization strategy on the GP posterior. \textbf{(a)}~Uniformly spread observations yield a well-calibrated posterior across the entire domain, enabling the surrogate to capture the global structure of the objective (dashed line). The true optimum (green dotted line) falls within a region of low uncertainty. \textbf{(b)}~Observations clustered near the library default produce a posterior that is overconfident locally but highly uncertain elsewhere, risking suppressed exploration near the true optimum.}
\label{fig:gp_prior_effect}
\end{figure}

The implications of the initialization strategy for the GP posterior are illustrated in Figure~\ref{fig:gp_prior_effect}. When initial observations are spread uniformly across the domain (panel~a), the GP posterior achieves low uncertainty throughout the space and can model the global structure of the objective, including distant regions that may contain the optimum. When initial observations are clustered near the default (panel~b), the posterior uncertainty collapses locally but remains high far from the cluster. This creates a risk of \emph{conformist} behavior: the GP becomes overly confident near the default and under-explores the rest of the space, potentially missing the global optimum. We note that this argument applies directly to GP-based surrogates (BoTorch, Scikit-Optimize); Optuna's TPE surrogate has a different exploration mechanism, though our empirical results show consistent behavior across all three back-ends. Whether this theoretical risk manifests in practice is an empirical question that our experiments are designed to answer.

\section{Related work}
\label{sec:related}

Bayesian optimization has been the focus of sustained research effort aimed at improving its performance and convergence rate, particularly in the context of machine-learning hyperparameter tuning. Contributions have targeted every component of the BO pipeline---acquisition functions, surrogate models, evaluation strategies, and the incorporation of prior knowledge---and the resulting landscape is both rich and rapidly evolving.

On the acquisition-function side, improvement-based rules such as Expected Improvement and its logarithmic variant LogEI provide strong empirical performance with modest computational overhead~\cite{ament2023unexpected}, while information-theoretic alternatives---Predictive Entropy Search~\cite{hernandez2014predictive} and Max-value Entropy Search~\cite{wang2017max}---explicitly target the reduction of uncertainty about the optimum and often require fewer evaluations to locate high-value regions. Trust-region methods such as TuRBO~\cite{eriksson2019scalable} address the challenge of high-dimensional search spaces by coupling global exploration with rigorous local exploitation, and batch BO schemes like Local Penalisation~\cite{gonzalez2016batch} issue multiple queries simultaneously to achieve near-linear wall-clock speed-ups on parallel hardware. Multi-fidelity BO~\cite{wu2020practical} further reduces cost by interleaving cheap, low-fidelity approximations of the objective with expensive high-fidelity evaluations, and tuning the surrogate kernel or acquisition-function hyperparameters themselves has been shown to bring additional efficiency gains~\cite{lindauer2019towards}.

The surrogate-modelling literature has likewise diversified well beyond classical stationary GP kernels. Deep-kernel learning~\cite{wilson2016deep} combines GPs with neural feature extractors to capture non-stationary structure without sacrificing calibrated uncertainty, yielding improved performance on complex vision and language benchmarks. When dimensionality is extreme, latent-space techniques such as REMBO~\cite{wang2016bayesian} embed the search in a random low-dimensional subspace, enabling BO to operate on hundreds or thousands of variables while retaining convergence guarantees provided the intrinsic dimension is low.

Beyond internal algorithmic refinements, a growing body of work demonstrates the value of incorporating external prior knowledge into the optimization loop. In robotics, user-specified priors on likely poses or controller parameters have been shown to shorten the learning curve for new manipulation tasks~\cite{mayr2022learning}. In the hyperparameter-tuning literature, the dominant paradigm for injecting prior information is transfer learning: GP surrogates augmented with task-similarity metrics or ensemble priors leverage traces from previous optimization campaigns~\cite{tighineanu2022transfer,bai2023transfer}, and meta-models that predict promising initial configurations can dramatically shorten subsequent BO runs~\cite{kim2017learning}. These approaches, however, require access to historical optimization data from related tasks, which may not always be available or easy to curate.

A natural and far simpler source of prior information---one that requires no historical data whatsoever---is the set of default hyperparameter values provided by mainstream ML libraries. These defaults have been selected by library maintainers through extensive empirical testing and encode pragmatic expert knowledge about broadly effective configurations. Because they are freely available and incur no additional cost, centering BO initialization around them could, in principle, offer a broadly applicable, zero-overhead improvement to existing tuning pipelines. Yet, to the best of our knowledge, no prior study has systematically evaluated whether these defaults constitute a useful initialization strategy for Bayesian optimization. The present work fills this gap by providing a rigorous, large-scale empirical assessment across multiple optimization back-ends, model families, and benchmark tasks.

\section{Research hypotheses and methodology}
\label{sec:method}

The central question of this study is whether the default hyperparameter configurations provided by standard machine-learning libraries can serve as effective informative initialization for Bayesian optimization. This section first states the research hypotheses formally, then describes the default-centered initialization strategy, and finally defines the evaluation metrics and statistical testing procedure used to assess them.

\paragraph{Research hypotheses}
We formulate two complementary hypotheses that capture the two dimensions along which default-centered initialization could provide an advantage: convergence speed and final solution quality. For each dimension, the null hypothesis $H_0$ states that the default-centered strategy performs no better than random initialization, and the alternative hypothesis $H_1$ states that it performs strictly better. Letting $p_{\mathrm{win}}$ denote the probability that the default-centered strategy outperforms the random baseline in a given comparison, the hypotheses are:

\medskip
\noindent\textbf{Hypothesis 1} (Convergence speed):
\begin{equation}
H_0^{\mathrm{conv}}\colon p_{\mathrm{win}}^{\mathrm{conv}} \leq 0.5 \qquad \text{vs.} \qquad H_1^{\mathrm{conv}}\colon p_{\mathrm{win}}^{\mathrm{conv}} > 0.5.
\label{eq:h_conv}
\end{equation}
Rejecting $H_0^{\mathrm{conv}}$ would indicate that initializing BO near the library defaults leads to systematically faster convergence than random initialization.

\medskip
\noindent\textbf{Hypothesis 2} (Final performance):
\begin{equation}
H_0^{\mathrm{metric}}\colon p_{\mathrm{win}}^{\mathrm{metric}} \leq 0.5 \qquad \text{vs.} \qquad H_1^{\mathrm{metric}}\colon p_{\mathrm{win}}^{\mathrm{metric}} > 0.5.
\label{eq:h_metric}
\end{equation}
Rejecting $H_0^{\mathrm{metric}}$ would indicate that default-centered initialization leads to systematically better final predictive performance.

\medskip
\noindent Both hypotheses are tested independently for each of the two default-centered strategies described below (Sample and Default), yielding four statistical tests in total.

\begin{figure}[t]
\centering
\includegraphics[width=0.75\textwidth]{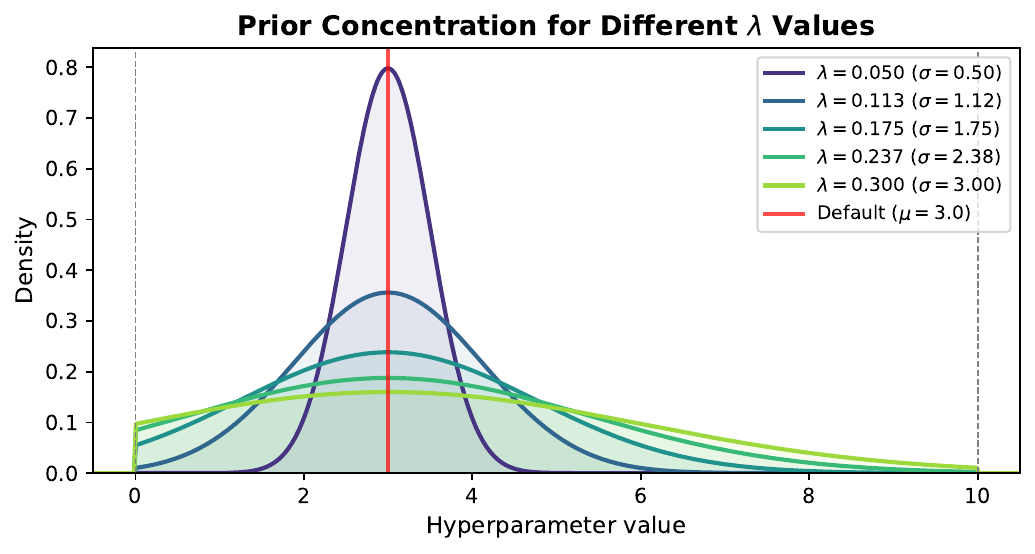}
\caption{Truncated Gaussian distributions for different values of the concentration parameter $\lambda$. For $\lambda = 0.05$, approximately 68\% of the probability mass lies within 10\% of the parameter range around the default; for $\lambda = 0.30$, the distribution is nearly uniform. The red vertical line marks the library default value.}
\label{fig:sensitivity_lambda}
\end{figure}

\paragraph{Default-centered initialization}
In our setting, the objective function $f\colon X \to \mathbb{R}$ is defined over a bounded hyperparameter domain $X \subset \mathbb{R}^d$, and $f(x)$ is the $k$-fold cross-validated performance of a machine-learning model trained with hyperparameters $x$:
\begin{equation}
f(x) = \frac{1}{k}\sum_{j=1}^{k} M_j(x),
\end{equation}
where $M_j(x)$ is the performance metric (accuracy for classification, negative root mean squared error for regression) on the $j$-th fold. Our proposed initialization replaces the standard uniform sampling with points drawn from a truncated normal distribution centered at the default hyperparameter values supplied by the ML library. Specifically, for a hyperparameter $x_i$ with default value $\mu_i$ and domain bounds $[a_i, b_i]$, the initial samples are drawn from
\begin{equation}
x_i \sim \mathrm{TruncNormal}\!\big(\mu_i,\, \sigma_i^{2};\, a_i,\, b_i\big),
\label{eq:truncnorm}
\end{equation}
where the standard deviation is set proportional to the range of the search space, $\sigma_i = \lambda\,(b_i - a_i)$, with $\lambda \in (0, 1)$ controlling the concentration of the initialization around the default. Small values of $\lambda$ produce a tightly concentrated distribution near the default, while larger values yield a diffuse distribution that approaches uniform sampling. Figure~\ref{fig:sensitivity_lambda} illustrates the effect of $\lambda$ on the shape of the truncated Gaussian: as $\lambda$ increases from 0.05 to 0.30, the distribution transitions from a narrow peak concentrated at the default to a broad, nearly uniform spread across the domain.

We evaluate two distinct initialization strategies against a common random baseline. In the \textbf{Default} strategy, BO is initialized with a single deterministic point corresponding to the library default configuration, and 15 subsequent BO iterations are performed. In the \textbf{Sample} strategy, BO is initialized with multiple points (three, four, or five) drawn from the truncated Gaussian of Eq.~(\ref{eq:truncnorm}), followed by 30 BO iterations. In both cases, the \textbf{Random} baseline uses the same total evaluation budget but draws all initial points from a uniform distribution over the domain.

\paragraph{Evaluation metrics}
For each experimental run, we record two quantities: the \emph{convergence index}, defined as the iteration at which the running-best objective value first reaches its final optimum, and the \emph{best metric}, defined as the highest objective value attained over the entire run. These are averaged across repetitions for each combination of optimization back-end, model, dataset, and initialization strategy. The relative improvement of the default-centered method over the random baseline is measured through
\begin{equation}
\Delta^{\mathrm{conv}}_{i,j} = \frac{\mu^{\mathrm{conv}}_{i,j}(\text{random}) - \mu^{\mathrm{conv}}_{i,j}(\text{default})}{\mu^{\mathrm{conv}}_{i,j}(\text{random})}, \qquad
\Delta^{\mathrm{metric}}_{i,j} = \frac{\mu^{\mathrm{metric}}_{i,j}(\text{default}) - \mu^{\mathrm{metric}}_{i,j}(\text{random})}{\mu^{\mathrm{metric}}_{i,j}(\text{random})},
\end{equation}
where $(i,j)$ indexes the dataset--model pair. A ``win'' for the default-centered strategy is recorded when $\Delta^{\mathrm{conv}}_{i,j}$ exceeds a convergence threshold $\tau_{\mathrm{conv}} = 0.10$ (a 10\% relative speed-up) or when $\Delta^{\mathrm{metric}}_{i,j}$ exceeds a metric threshold $\tau_{\mathrm{metric}}$ set to the minimum relative spread observed across datasets (0.3\%). Ties---cases where neither strategy exceeds the threshold---are excluded from the statistical test.

\paragraph{Statistical testing}
Each hypothesis is tested via a one-sided binomial test. Let $N_{\mathrm{win}}$ be the number of dataset--model--library combinations in which the default-centered strategy wins, and let $N_{\mathrm{total}}$ be the total number of non-tied comparisons. Under $H_0$, the number of wins follows $N_{\mathrm{win}} \sim \mathrm{Binomial}(N_{\mathrm{total}},\, 0.5)$, and the $p$-value is $P(N_{\mathrm{win}} \geq n_{\mathrm{obs}} \mid p = 0.5)$. Rejecting $H_0$ at significance level $\alpha = 0.05$ requires the $p$-value to fall below 0.05, providing evidence that the default-centered strategy is systematically superior. We apply this test separately for each of the four comparisons: Sample vs.\ Random on convergence (Hypothesis~1), Sample vs.\ Random on metric (Hypothesis~2), Default vs.\ Random on convergence (Hypothesis~1), and Default vs.\ Random on metric (Hypothesis~2).

We note that failure to reject $H_0$ does not prove that default-centered initialization is exactly as effective as random initialization---it establishes that the available evidence is insufficient to conclude it is better. However, the consistency of the non-significant results across all four comparisons, multiple models, datasets, and BO back-ends provides strong practical evidence that any benefit, if it exists, is too small to be of practical relevance.

\section{Experiments}
\label{sec:experiments}

The experimental design is governed by the need to ensure that conclusions generalize across the heterogeneous landscape of machine-learning models, optimization back-ends, and data characteristics. We therefore construct a full factorial benchmark that spans three orthogonal axes: the choice of BO library, the choice of ML model family, and the choice of dataset.

On the optimization side, three libraries are employed to cover the main families of surrogate models used in practice. Scikit-Optimize (Skopt) provides a GP-based optimizer with a simple interface, BoTorch offers a more advanced GP framework powered by gradient-based acquisition-function optimization within the PyTorch ecosystem, and Optuna implements the Tree-structured Parzen Estimator (TPE), a tree-based nonparametric surrogate. By including both kernel-based probabilistic surrogates (BoTorch, Skopt) and tree-based nonparametric alternatives (Optuna), the benchmark avoids conclusions that might be an artifact of a single surrogate architecture.

The model families under study are Random Forests, Support Vector Machines (SVC for classification, SVR for regression), and Multilayer Perceptrons (MLP), all drawn from the scikit-learn library~\cite{pedregosa2011scikit}. These three families were selected because they represent fundamentally different inductive biases---ensemble-based decision trees, margin-based kernel methods, and deep nonlinear function approximators---and because each exposes a moderately sized but non-trivial hyperparameter space. Linear models such as logistic regression were excluded because their hyperparameter spaces are too small to benefit from, or meaningfully stress, an optimization procedure. The default values used in our experiments are based on the scikit-learn library defaults and established recommendations from the machine-learning literature (e.g., Breiman's recommendations for Random Forest regressors). The complete list of defaults, search bounds, and descriptions is provided in the Appendix (Tables~\ref{tab:hp_rf},~\ref{tab:hp_mlp}, and~\ref{tab:hp_svm}).

Five benchmark datasets are used, three for classification and two for regression, as summarized in Table~\ref{tab:datasets}. The datasets are chosen to cover a range of domain characteristics including varying feature dimensionality, class balance, noise levels, and sensitivity to hyperparameter choice. To control training time and ensure reproducibility, all datasets are subsampled to a maximum of 1{,}000 instances using a fixed random seed. We acknowledge that this subsampling may alter the characteristics of originally large-scale datasets (e.g., Higgs, Airlines10M); this trade-off was necessary to enable the thousands of model evaluations required by the full factorial design.

\begin{table}[t]
\centering
\caption{Benchmark datasets used in the experimental evaluation.}
\label{tab:datasets}
\begin{tabular}{llll}
\toprule
Name & Task & Description \\
\midrule
AmazonAcc   & Classification & Binary access prediction from human resources \\
Letter      & Classification & Multiclass character recognition from images \\
Higgs       & Classification & Particle physics dataset (large and noisy) \\
BikeSharing & Regression     & Bike rental demand over time \\
Airlines10M & Regression     & Airline delay prediction (very large-scale) \\
\bottomrule
\end{tabular}
\end{table}

All experiments use 3-fold cross-validation ($k=3$) to estimate the performance of each hyperparameter configuration. The Sample strategy is tested with three, four, and five initialization points and five evenly spaced values of the prior-concentration parameter $\lambda \in \{0.05, 0.1125, 0.175, 0.2375, 0.30\}$. For $\lambda = 0.05$, approximately 68\% of the probability mass lies within $\pm\sigma$ of the default, corresponding to only 10\% of the total parameter range; for $\lambda = 0.30$, the distribution is substantially more diffuse and approaches a near-uniform spread. Each configuration is repeated multiple times to enable meaningful averaging, yielding a total of 48 non-trivial dataset--model--library combinations for the main binomial analysis.\footnote{The 48 combinations arise from the full factorial crossing of datasets, models, libraries, and initialization-point counts, after aggregation over $\lambda$ values and repetitions.} The hardware environment consists of a Linux workstation with 60~GB of RAM and an NVIDIA RTX 3060 GPU with 12~GB of memory.

\section{Results and discussion}
\label{sec:results}

We organize the presentation of results in three stages: a qualitative visual inspection of convergence behavior, a quantitative statistical analysis based on the binomial test, and a sensitivity study that probes the effect of the prior variance on optimization dynamics.

To guide the qualitative analysis, we first identify the datasets on which hyperparameter configuration has the greatest impact by computing the \emph{spread} metric for each dataset $D_i$,
\begin{equation}
\mathrm{Spread}(D_i) = \frac{1}{|M|\,|S|}\sum_{(j,s)\in M\times S} \frac{\max_t r^{(s)}_{i,j}(t) - \min_t r^{(s)}_{i,j}(t)}{\min_t r^{(s)}_{i,j}(t)},
\label{eq:spread}
\end{equation}
where $r^{(s)}_{i,j}(t)$ is the running-best metric at iteration $t$ for model $j$ under strategy $s$. The spread values, reported in Table~\ref{tab:spread}, range from 0.003 (AmazonAcc) to 0.18 (BikeSharing), indicating that some datasets are far more sensitive to hyperparameter choice than others. We focus the visual analysis on the high-spread datasets (BikeSharing, Letter) where differences between strategies, if they exist, should be most visible.

\begin{table}[t]
\centering
\caption{Spread distance (Eq.~\ref{eq:spread}) for each dataset. Higher values indicate greater sensitivity to hyperparameter configuration.}
\label{tab:spread}
\begin{tabular}{lc}
\toprule
Dataset & Spread \\
\midrule
Airlines10M & 0.019 \\
AmazonAcc   & 0.003 \\
BikeSharing & 0.180 \\
Higgs       & 0.017 \\
Letter      & 0.120 \\
\bottomrule
\end{tabular}
\end{table}

The distributional summaries of final performance, displayed as box plots normalized within each dataset, reveal no systematic or practically meaningful difference between the competing initialization strategies. After normalization, the inter-quartile ranges of the default-initialized runs (15-evaluation budget) and the sample-initialized runs overlap almost completely with those of the random baseline (30-evaluation budget). No consistent shift in central tendency or dispersion is observed. The convergence trajectories corroborate this finding: while the sample-initialized curves occasionally start at a higher running-best value than their random counterparts---a natural consequence of sampling near defaults that are engineered to be safe configurations---the gap contracts rapidly, and both strategies approach a common plateau well within the allotted evaluation budget. When the budget is restricted to 15 evaluations, even the initial separation is negligible, and the two methods stabilize at indistinguishable performance levels. Whatever exploratory head start the default-centered initialization may enjoy at iteration zero is therefore transient; random search closes the gap quickly, and all strategies converge to essentially the same asymptotic outcome.

The quantitative analysis confirms these visual impressions with formal statistical evidence. Table~\ref{tab:binomial} reports the win counts, tie counts, and one-sided binomial $p$-values for each of the four comparisons.

\begin{table}[t]
\centering
\caption{Summary of win counts, ties, and one-sided binomial $p$-values for each comparison. S~refers to the Sample strategy, R~to Random, and D~to Default. Wins for S/D are counted when the default-centered method exceeds the threshold; wins for R are counted in the reverse direction. None of the $p$-values approaches the 0.05 significance level, and therefore we fail to reject any of the four null hypotheses.}
\label{tab:binomial}
\begin{tabular}{lccccl}
\toprule
Comparison & S/D & R & Ties & $p$-value & Decision \\
\midrule
S vs.\ R Convergence  & 18 & 15 & 15 & 0.364 & Fail to reject $H_0^{\mathrm{conv}}$ \\
S vs.\ R Metric       & 10 & 12 & 26 & 0.738 & Fail to reject $H_0^{\mathrm{metric}}$ \\
D vs.\ R Convergence  & 19 & 12 & 17 & 0.141 & Fail to reject $H_0^{\mathrm{conv}}$ \\
D vs.\ R Metric       & 11 & 17 & 20 & 0.908 & Fail to reject $H_0^{\mathrm{metric}}$ \\
\bottomrule
\end{tabular}
\end{table}

For convergence speed under the Sample strategy, the default-centered method wins in 18 out of 48 combinations while the random baseline wins in 15, with 15 ties; the resulting $p$-value of 0.364 is far from significance. On the metric side for 30 evaluations, the sample-centered method wins in only 10 cases against 12 for random (26 ties), yielding $p = 0.738$. Moving to the 15-evaluation comparison, the Default strategy converges at least 10\% faster than random in 19 cases versus 12 for random ($p = 0.141$), and secures a metric advantage in 11 cases versus 17 for random ($p = 0.908$). None of the four $p$-values comes close to the conventional $\alpha = 0.05$ threshold, and the metric comparison for the Default strategy in fact trends in the wrong direction, with random winning more often than default. Taken together, these results provide no evidence to reject any of the null hypotheses stated in Eqs.~(\ref{eq:h_conv})--(\ref{eq:h_metric}): neither the sample-based approach nor the deterministic-default initialization produces a statistically significant advantage over purely random initialization.

The sensitivity analysis examines how the concentration parameter $\lambda$ of the truncated Gaussian affects optimization dynamics, focusing on the Letter dataset, which was identified as particularly sensitive to hyperparameter configuration by the spread metric. We track three normalized metrics across $\lambda$ values: the maximum performance attained, the mean running-best performance, and the convergence index. All metrics are averaged across models and normalized via min--max scaling for visual comparability.

The maximum (final) performance remains essentially flat across all values of $\lambda$ (Pearson $r = -0.182$, $p = 0.335$), confirming that the optimizer consistently reaches high-quality solutions regardless of the initialization width. However, the mean running-best performance exhibits a statistically significant negative correlation with $\lambda$ (Pearson $r = -0.626$, $p = 2.185 \times 10^{-4}$), indicating that broader initialization distributions degrade the average quality of configurations evaluated throughout the run. This degradation is not caused by slower convergence---the convergence index shows no significant correlation with $\lambda$ ($r = 0.048$, $p = 0.801$)---but rather by poorer early evaluations. A decomposition into early-phase and late-phase mean performance confirms this interpretation: the early mean decreases sharply with $\lambda$ ($r = -0.641$, $p = 1.368 \times 10^{-4}$), while the late mean remains stable ($r = -0.185$, $p = 0.328$).

\begin{table}[t]
\centering
\caption{Pearson correlation coefficients between the initialization concentration parameter ($\lambda$) and various normalized metrics on the Letter dataset.}
\label{tab:sensitivity}
\begin{tabular}{lcc}
\toprule
Metric & Pearson $r$ & $p$-value \\
\midrule
Max.\ performance   & $-0.182$ & $0.335$ \\
Mean performance     & $-0.626$ & $2.185 \times 10^{-4}$ \\
Convergence index    & $\phantom{-}0.048$ & $0.801$ \\
Early mean           & $-0.641$ & $1.368 \times 10^{-4}$ \\
Late mean            & $-0.185$ & $0.328$ \\
\bottomrule
\end{tabular}
\end{table}

These findings illuminate the mechanism behind the superficial appeal of default-centered initialization. Default hyperparameters are deliberately chosen by library maintainers to lie in ``safe'' high-performing regions of the search space, so evaluating configurations near them during the early iterations naturally yields decent scores. This gives the impression of a useful head start. However, this initial advantage carries no directional information for the optimizer: the defaults do not indicate where the global optimum lies, nor do they reveal the shape of the loss landscape. As illustrated in Figure~\ref{fig:gp_prior_effect}, the GP surrogate gains no exploitable structure from a cluster of similarly performing initial points near the default; the posterior simply becomes overconfident locally while remaining uninformed about the rest of the domain. A uniformly sampled set of initial points, by contrast, provides broader coverage of the search space and gives the surrogate a more informative basis for modeling the objective surface. The result is that BO quickly compensates for any early disadvantage of random initialization, and the two strategies converge to indistinguishable outcomes.

\paragraph{Limitations}
Several aspects of the experimental design warrant discussion. First, the sensitivity analysis is restricted to the Letter dataset; while this was the most informative dataset by the spread metric, extending the analysis to additional datasets would strengthen the conclusions. Second, the statistical power of the binomial test is limited by the number of non-tied comparisons (as few as 22 for S vs.\ R Metric); larger-scale experiments could reveal small but statistically significant effects that the current study cannot detect. Third, using $k = 3$ folds for cross-validation, while computationally practical, introduces more variance in performance estimates than $k = 5$ or $k = 10$. Finally, all datasets are subsampled to 1{,}000 instances, which may alter the relationship between hyperparameters and performance for originally large-scale datasets.

\section{Conclusions and future work}
\label{sec:conclusion}

This study set out to determine whether the default hyperparameter configurations provided by popular machine-learning libraries can serve as effective informative initialization for Bayesian optimization. The hypothesis is intuitively appealing: if defaults encode expert knowledge about broadly effective configurations, centering the BO initialization around them should accelerate convergence and improve final performance relative to uninformed random search. Our extensive empirical evaluation, spanning three BO back-ends, three model families, five benchmark datasets, and multiple prior-variance settings, provides a clear and statistically grounded answer: we find no evidence that default hyperparameters provide informative initialization for BO.

None of the four statistical tests---covering both convergence speed and final metric quality under both deterministic-default and probabilistic-sampling initialization---produced a $p$-value below 0.14, far from any conventional significance threshold. The null hypotheses $H_0^{\mathrm{conv}}$ and $H_0^{\mathrm{metric}}$ could not be rejected in any of the four comparisons, meaning there is no statistical evidence that default-centered initialization outperforms random initialization. The sensitivity analysis further revealed that the apparent benefit of default-centered initialization is limited to the earliest iterations, where configurations near the safe default naturally score well, but this transient advantage is rapidly erased as BO explores the space. The final performance is invariant to the initialization width, confirming that the optimizer's own exploration mechanism, rather than the initial configuration, determines the quality of the solution.

These findings carry a direct and actionable message for practitioners. Default hyperparameters are convenient starting points for rapid prototyping and single-shot deployment, but they should not be mistaken for informative guides to the loss landscape. Relying on defaults when performance is a requirement may encourage premature convergence or instill a false sense of optimality, ultimately hindering rather than helping the search. Bayesian optimization and other principled search strategies remain essential precisely because they provide structured, uncertainty-aware exploration of the hyperparameter space based on actual performance feedback, not on assumptions rooted in library defaults.

Several directions for future work emerge naturally from these conclusions. First, rather than relying on static defaults, meta-learning frameworks that learn task-specific priors from historical optimization traces~\cite{bai2023transfer,tighineanu2022transfer} represent a more principled source of prior information, and a comparative study between meta-learned priors and default-centered initialization would quantify the gap. Second, extending the analysis to deep-learning models---including fine-tuning of large pre-trained architectures---would test whether the same negative result holds in settings where the hyperparameter space is larger and the evaluation cost is substantially higher. Third, the current study uses Expected Improvement as the sole acquisition function; evaluating the interaction between the initialization strategy and alternative acquisition functions (e.g., Knowledge Gradient, Thompson Sampling) could reveal settings in which the initialization matters more or less. Finally, investigating multi-fidelity BO methods that can leverage low-cost early-stopping evaluations to guide hyperparameter search more efficiently may offer a more productive avenue for reducing the cost of tuning than attempting to extract prior information from defaults.

\section*{Acknowledgments}
This work was partially supported by Universidad Pontificia Comillas and the Institute for Research in Technology (IIT). The authors thank Sim\'on Rodr\'iguez Santana for co-directing the thesis from which this work originates.

\bibliographystyle{elsarticle-num}

\appendix
\section{Hyperparameter configurations}
\label{app:hyperparams}

The default values listed below are based on scikit-learn library defaults and established recommendations from the machine-learning literature~\cite{pedregosa2011scikit}. For Random Forest regressors, the \texttt{max\_features} and \texttt{min\_samples\_leaf} defaults follow Breiman's original recommendations rather than the scikit-learn programmatic defaults. For SVM, the \texttt{gamma} value is a fixed approximation used in place of scikit-learn's data-dependent \texttt{'scale'} default (which computes $1/(d \cdot \mathrm{Var}(X))$) to enable a consistent initialization across datasets.

\begin{table}[ht]
\centering
\caption{Random Forest hyperparameter defaults (values are \textit{Classifier}, \textit{Regressor}), search bounds, and descriptions.}
\label{tab:hp_rf}
\begin{tabular}{llll}
\toprule
Hyperparameter & Default (C, R) & Bounds & Description \\
\midrule
n\_estimators  & 500, 500 & $[100, 2000]$ & Number of trees \\
max\_features   & $\sqrt{d}$, $d/3$ & $[1, d]$ & Features per split \\
max\_samples    & all, all & $[0.5n, n]$ & Samples per tree \\
min\_samples\_leaf & 1, 5 & $[1, 20]$ (C), $[2, 50]$ (R) & Min.\ leaf samples \\
\bottomrule
\end{tabular}
\end{table}

\begin{table}[ht]
\centering
\caption{Multilayer Perceptron hyperparameter defaults, search bounds, and descriptions.}
\label{tab:hp_mlp}
\begin{tabular}{llll}
\toprule
Hyperparameter & Default & Bounds & Description \\
\midrule
n\_layers      & 1       & $[1, 5]$             & Hidden layers \\
hidden\_units  & 100     & $[50, 500]$          & Units per layer \\
alpha          & 0.001   & $[10^{-5}, 10^{-1}]$ & $L_2$ regularization \\
lr             & 0.01    & $[10^{-4}, 10^{-1}]$ & Learning rate \\
\bottomrule
\end{tabular}
\end{table}

\begin{table}[ht]
\centering
\caption{Support Vector Machine hyperparameter defaults, search bounds, and descriptions.}
\label{tab:hp_svm}
\begin{tabular}{lllll}
\toprule
Hyperparameter & Default & Bounds & Applies to & Description \\
\midrule
C        & 1.0     & $[0.1, 10.0]$         & SVC, SVR & Regularization \\
gamma    & 0.01    & $[10^{-4}, 1.0]$      & SVC, SVR & Kernel coefficient \\
degree   & 3       & $[2, 5]$              & SVC, SVR & Polynomial degree \\
coef0    & 0.0     & $[0.0, 1.0]$          & SVC only & Polynomial offset \\
tol      & $10^{-3}$ & $[10^{-4}, 10^{-1}]$ & SVC only & Stopping tolerance \\
epsilon  & 0.1     & $[10^{-4}, 1.0]$      & SVR only & $\varepsilon$-insensitive tube \\
\bottomrule
\end{tabular}
\end{table}

\end{document}